\documentclass[journal]{IEEEtran}
\usepackage{cite}

\ifCLASSINFOpdf
  \usepackage[pdftex]{graphicx}
  \graphicspath{{./images/}}
  \DeclareGraphicsExtensions{.pdf}
\else
\fi
\usepackage{amsmath}
\usepackage{algpseudocode}
\usepackage{algorithm}
\ifCLASSOPTIONcompsoc
 \usepackage[caption=false,font=normalsize,labelfont=sf,textfont=sf]{subfig}
\else
 \usepackage[caption=false,font=footnotesize]{subfig}
\fi
\usepackage{url}

\usepackage[hidelinks]{hyperref}
\usepackage[utf8]{inputenc}

\usepackage{amssymb} 
\usepackage{array} 
\usepackage{booktabs} 
\usepackage{multirow}
\usepackage{tikz} 
\usetikzlibrary{shapes.geometric, arrows, calc}
\usepackage[capitalise]{cleveref} 
\usepackage{enumitem} 
\usepackage[switch]{lineno}  

\newcommand{\MDP}{\mathcal{M}} 
\newcommand{\Acts}{\mathcal{A}} 
\newcommand{\Trans}{T} 
\newcommand{\Rewards}{R} 
\newcommand{\States}{\mathcal{S}} 

\newcommand{\Task}{T} 
\newcommand{\Domain}{D} 
\newcommand{\Instance}{N} 
\newcommand{\Types}{\mathcal{T}} 
\newcommand{\Preds}{\mathcal{P}} 
\newcommand{\Funcs}{\mathcal{F}} 
\newcommand{\ActsP}{\mathcal{O}} 
\newcommand{\Objects}{B} 
\newcommand{\Init}{I} 
\newcommand{\Goal}{G} 

\newcommand{\RR}{\mathbb{R}} 

\algnewcommand{\algorithmicgoto}{\textbf{go to}}%
\algnewcommand{\Goto}[1]{\algorithmicgoto~\ref{#1}}%

\tikzstyle{module} = [rectangle, rounded corners, minimum width=1.8cm, minimum height=0.8cm,text centered, text width=2cm, draw=black]
\tikzstyle{representation} = [diamond, minimum width=1.6cm, minimum height=1.6cm, text centered, text width=1.5cm, draw=black, inner sep=-3pt]
\tikzstyle{invis} = [draw=none]
\tikzstyle{action} = [rectangle, minimum width=1cm, minimum height=0.6cm,text centered, text width=1cm, draw=black]
\tikzstyle{zone} = [rectangle, fill=black!20, draw=black!30, fill opacity=0.2]
\tikzstyle{label} = [rectangle, rounded corners,anchor=west,inner sep=2pt,minimum width=4ex, draw=black!30,fill=white]
\tikzstyle{observation} = [circle, inner sep=2pt,minimum width=0.2cm,minimum height=0.2cm,draw=black]
\tikzstyle{box} = [rectangle, minimum width=2cm, minimum height=0.6cm,text centered, draw=black]

\tikzstyle{environment} = [ellipse, minimum width=1cm, minimum height=3.6cm, text centered, draw=black]
\tikzstyle{environment2} = [ellipse, minimum width=2cm, minimum height=0.6cm, text centered, draw=black]
\tikzstyle{module2} = [rectangle, rounded corners, minimum width=2cm, minimum height=1cm,text centered, text width=2cm, draw=black]
\tikzstyle{arrow} = [thick,->,>=stealth]
\tikzstyle{rarrow} = [thick,<-,>=stealth]
\tikzstyle{dashedarrow} = [thick,->,>=stealth,dashed]
\tikzstyle{rdashedarrow} = [thick,<-,>=stealth,dashed]
\tikzstyle{dottedarrow} = [thick,->,>=stealth,dotted]
\tikzstyle{rdottedarrow} = [thick,<-,>=stealth,dotted]
\tikzstyle{line} = [thick,-]
\tikzstyle{dottedline} = [thick,-,dotted]

\newcolumntype{!}{>{\global\let\currentrowstyle\relax}}
\newcolumntype{^}{>{\currentrowstyle}}
\newcommand{\rowstyle}[1]{\gdef\currentrowstyle{#1}%
  #1\ignorespaces
}

\begin{document}
%
\title{SAGE: Generating Symbolic Goals for Myopic Models in Deep Reinforcement Learning}
%
%
%

\author{Andrew~Chester, 
        Michael~Dann, 
        Fabio~Zambetta, 
        and~John~Thangarajah
    \thanks{Manuscript submitted May 28, 2021.}
    \thanks{All authors affiliated with the School of Computing Technologies, RMIT University, Melbourne, Australia.
            Corresponding author Andrew Chester: andrew.chester@rmit.edu.au}
    \thanks{© 2021 IEEE. Personal use of this material is permitted. Permission
    from IEEE must be obtained for all other uses, in any current or future
    media, including reprinting/republishing this material for advertising or
    promotional purposes, creating new collective works, for resale or
    redistribution to servers or lists, or reuse of any copyrighted
    component of this work in other works.}
    }

%
%

\markboth{Journal of \LaTeX\ Class Files,~Vol.~14, No.~8, August~2015}%
{Shell \MakeLowercase{\textit{et al.}}: Bare Demo of IEEEtran.cls for IEEE Journals}
%



\maketitle

\begin{abstract}
Model-based reinforcement learning algorithms are typically more sample efficient than their model-free counterparts, especially in sparse reward problems.
Unfortunately, many interesting domains are too complex to specify the complete models required by traditional model-based approaches.
Learning a model takes a large number of environment samples, and may not capture critical information if the environment is hard to explore.
If we could specify an incomplete model and allow the agent to learn how best to use it, we could take advantage of our partial understanding of many domains. 
Existing hybrid planning and learning systems which address this problem often impose highly restrictive assumptions on the sorts of models which can be used, limiting their applicability to a wide range of domains.
In this work we propose SAGE, an algorithm combining learning and planning to exploit a previously unusable class of incomplete models. 
This combines the strengths of symbolic planning and neural learning approaches in a novel way that outperforms competing methods on variations of taxi world and Minecraft.
\end{abstract}

\begin{IEEEkeywords}
Reinforcement Learning, Deep Learning, Neural Networks, Symbolic Planning, Neuro-Symbolic
\end{IEEEkeywords}

%
\IEEEpeerreviewmaketitle

\section{Introduction}
%
%
%
%

\IEEEPARstart{D}{eep} reinforcement learning (DRL) agents have shown remarkable progress in a range of application domains, from video games \cite{mnihHumanlevelControlDeep2015} to robotic control \cite{haarnojaSoftActorCriticOffPolicy2018}, but still lag far behind humans in their ability to learn quickly from limited experience.
Some of the best performing DRL agents are in board games, largely due to the fact that these agents are provided with a full deterministic model of their environment \cite{silverGeneralReinforcementLearning2018}. 
Model-based reinforcement learning allows agents to plan ahead, which can guide exploration efficiently. 
In most real world domains it is impractical (if not impossible) to provide a complete model of the environmental dynamics,  and despite some promising recent work \cite{schrittwieserMasteringAtariGo2020}, learning a model is often highly sample inefficient.
However, in many cases it is feasible to specify an incomplete model which captures only some aspects of the domain. 

\begin{figure}
    \centering
    \begin{tikzpicture}[node distance=1cm]

        \node (ma) [box, minimum width=2.9cm]  at (0,0.6) {Myopic \& abstract};
        \node (m) [box, minimum width=5.65cm]  at (-1.375,1.4) {Myopic};
        \node (a) [box, minimum width=5.65cm]  at (1.375,2.2) {Abstract};
        \node (c) [box, minimum width=8.4cm]  at (0,3) {Complete};

        \draw [arrow] 
                    (-4.2,0) node[anchor=south west] {short term} 
                             node[anchor=north west,shift={(0,-0.45)}, text width=2.4cm] {Low-level vehicle control} 
                    -- node[anchor=north, text width=2cm] {\textbf{Examples:}\\Route navigation}  
                    (4.2,0) node[anchor=south east] {long term} 
                            node[anchor=north east,shift={(0,-0.45)}, text width=1.8cm] {Passenger locations};
         
    \end{tikzpicture}
    \caption{Different types of models classified according to their suitability for use in long and short term planning. Examples show the information needed to plan over that timescale in the taxi domain.}
\label{fig:models}
\end{figure}
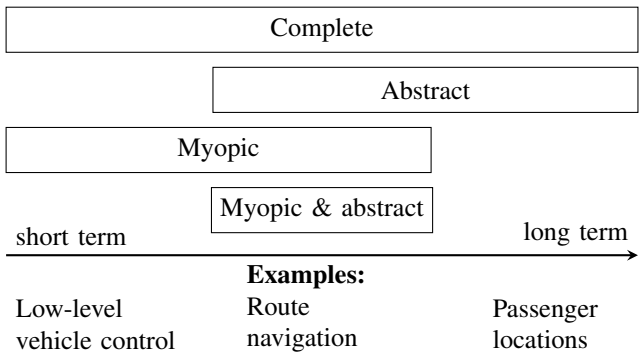

The key contribution of this paper is a novel integrated planning and learning algorithm which enables the use of incomplete symbolic domain models that cannot be exploited by prior work, providing more efficient learning in challenging environments.
More specifically, it allows agents to use models that are simultaneously  \emph{abstract} -- meaning the model cannot be used for short term planning -- and \emph{myopic} -- meaning the model cannot be used for long term planning (\cref{fig:models}).

To illustrate the concepts of myopic and abstract models, we will use a taxi domain modelled after the task facing a rideshare driver.
Consider a single taxi operating in a known road network. 
Passengers appear at random times and locations according to an unknown distribution, and have random destinations drawn from a separate unknown distribution. 
The taxi driver has access to a map, which shows the location and destination of any currently available passengers. 
The taxi can only pick up one passenger at a time, and the goal of the taxi driver is to maximise the number of passengers delivered in a fixed time period. 

A human faced with this task is likely to approach it hierarchically, deciding on long term goals before filling in the details. 
First, they decide what they want to accomplish at a high level: ``should I pick up passenger A, B, or refuel"?
Then, they would plan out a route to navigate to their intended destination: ``To get to that passenger I will need to take the next right and then go straight for three kilometres."
Finally, in the short term they control the car by steering, braking, etc.
In this example it is relatively easy to define an abstract and myopic model that is appropriate for mid-range planning, i.e. how to navigate through the streets.
However, defining a model that would allow us to maximise daily income (long term) would require, at minimum, an understanding of where passengers are likely to be at different times of day, and controlling the vehicle at the low-level (short term) would need a full physics simulator. 

Abstract planning models often do not even contain the objects that would be needed to perform short term planning; for example, a navigation model has no concept of a steering wheel. 
Likewise, myopic models lack some of the requisite concepts entirely, for example a navigation model with no concept of money cannot plan to earn the most cash by the end of the day.
Myopic models may also be deficient in other ways, such as not capturing stochasticity, or partial observability.
These inaccuracies in the model will accumulate over the lifetime of the plan, making long term predictions futile, but they can still be used effectively for mid-range planning. 

In theory, a hierarchical system could exploit such incomplete models by using model-free RL to learn control policies for the short and/or long term.
To this end, we propose SAGE (Symbolic Ancillary Goal gEneration), a three-tier architecture consisting of: 
\begin{itemize}
    \item a meta-controller that sets symbolic goals,
    \item a planner which produces a sequence of symbolic steps to attain the chosen goal, and
    \item a low-level controller which uses these steps as guidance for atomic actions.
\end{itemize}

SAGE's top-level meta-controller uses RL to set \emph{ancillary} symbolic goals. 
Ancillary goals are not defined in terms of the primary objective for the agent (maximising reward), as the myopic model may not capture expected reward. 
Instead, they are used to guide the lower layers of the hierarchy, allowing the meta-controller to focus on \emph{what} should be done, as opposed to \emph{how} it should be done.

Compared to other recent methods which blend RL and symbolic planning \cite{yangPEORLIntegratingSymbolic2018, lyuSDRLInterpretableDataefficient2019,gordonWhatShouldNow2019} our approach is novel in a number of ways.
We provide the first system that allows the meta-controller to set arbitrary goals in the symbolic space, instead of being restricted to a simple numerical threshold or planning according to a predetermined goal.
SAGE can also make use of models that are both myopic and abstract, unlike prior work.
This provides more flexibility for the model designer, reducing the manual effort required.

We demonstrate that SAGE learns more efficiently than competing approaches in a small taxi domain, and continues to learn near-optimal policies as the environment scales up in size, unlike our benchmarks. 
We also show that the planning component is crucial to its success by comparing against an ablated hierarchical method on a complex CraftWorld task.

\section{Background}

\subsection{Reinforcement learning}
We now briefly review the notation we will be using in this paper. A Markov Decision Process (MDP) $\MDP$ consists of a tuple $\MDP = (\States,\Acts,\Trans,\Rewards)$, where $s \in \States$ are the possible environment states, $a \in \Acts$ are the allowable actions, $\Trans : (\States\times\Acts\times\States)\rightarrow[0,1]$ is the transition matrix specifying the environment transition probabilities, and $\Rewards : (\States\times\Acts\times\mathcal{R})\rightarrow[0,1]$ specifies the reward probabilities ($\mathcal{R} \subset \RR$ is the range of allowable rewards). 
The objective of the agent is to maximise the total \emph{return} $U$, which is exponentially discounted by a \emph{discount rate} $\gamma \in [0,1]$, that is $U_t = \sum^{T}_{k=0}\gamma^k r_{t+k+1}$, where $T$ denotes the length of the episode remaining.
The agent takes actions according to some \emph{policy} $\pi(a|s) : (\States\times\Acts) \rightarrow [0,1]$, which defines the probability of taking each action in each state.
We denote sampling of an action according to the policy by $a_t \sim \pi(\cdot | s_t)$.

\subsection{Symbolic planning}
In this work we use the planning domain definition language (PDDL) to describe symbolic planning problems \cite{mcdermottPDDLPlanningDomain1998}. 
Formally, a planning task $\Task$ is a pair: $(\Domain,\Instance)$, where $\Domain$ is the general planning \emph{domain}, and $\Instance$ is the particular \emph{instance} to be solved. The domain defines the tuple $(\Types,\Preds,\Funcs,\ActsP)$, where $\Types$ is a set of types for objects, $\Preds$ is a set of Boolean predicates of the form $P(b_1,\dots,b_k)$, and $\Funcs$ is a set of numeric functions of the form $f(b_1,\dots,b_k)$, where $b_1,\dots,b_k$ are object variables. 
$\ActsP$ is the set of planning operators (referred to throughout as \emph{symbolic steps}), which have \emph{preconditions} (Boolean formulas over predicates and functions which must hold for the operator to be valid) and \emph{postconditions} (assignments to predicates or functions which are changed by the operator).
The instance contains $(\Objects,\Init,\Goal)$, where $\Objects$ are definitions of the objects, $\Init$ is a conjunction of predicates (or their negations) and functions that describe the initial state of the world, and $\Goal$ is the goal which likewise describes the desired end-state of the world.
The goal may only partially constrain the state of the world by omitting predicates and functions that may take any value. 

\section{Related work}

\subsection{Hierarchical RL}
Hierarchical reinforcement learning (HRL) agents consist of a number of layers, each acting at a higher level of abstraction than the last. 
Given suitable abstractions, this approach improves sample efficiency in sparse reward learning tasks, primarily due to temporally extended and semantically consistent exploration \cite{nachumWhyDoesHierarchy2019}.
The most common formalism for HRL is the options framework \cite{suttonMDPsSemiMDPsFramework1999}, in which the top layer chooses from a set of \emph{options}, rather than the base actions in the original MDP.
Each option is implemented as a different policy by the lower layer of the agent, and executes until a termination condition is reached.
There are a variety of different ways to extend HRL with deep learning \cite{vezhnevetsFeUdalNetworksHierarchical2017,nachumDataEfficientHierarchicalReinforcement2018,levyLearningMultilevelHierarchies2019,liSubpolicyAdaptationHierarchical2020,bagariaOptionDiscoveryUsing2020}.
Hierarchical DQN (hDQN) \cite{kulkarniHierarchicalDeepReinforcement2016} is most relevant to our work as it allows manually specified options, similar to our provided planning model.
Unlike our work, these hierarchical approaches cannot use a provided planning model to reduce sample complexity.

\begin{table*}[t!]
    \caption{Related Work Summary. Architecture column symbols are: M: meta-controller, P: planner, C: controller, arranged in hierarchical order (descending). C $\sim$ C implies a variable-tiered hierarchy of controllers. All methods require a provided state abstraction mapping, although Winder et al. 2020 can alternatively use one learned from labelled examples using HierGen \protect \cite{mehtaHierarchicalStructureDiscovery2011}}
    \label{tab:relatedwork}
    \begin{center}
        \small
        \begin{tabular}{!l^l^l^p{1.6cm}^l^l^l}
            \rowstyle{\bfseries}
            Authors & Architecture &  Planner type & Model type & Plan goal & Meta-controller & Controller \\\hline
            Roderick et al. 2018 \cite{roderickDeepAbstractQnetworks2018} & M C & MDP (Value iter) & N/A & Implicit & Learned (tabular) & Learned (deep)\\ 
            Gopalan et al. 2017 \cite{gopalanPlanningAbstractMarkov2017} & C $\sim$ C & MDP (Value iter) & N/A & Implicit & N/A & Provided (tabular)\\ 
            Winder et al. 2020 \cite{winderPlanningAbstractLearned2020} & C $\sim$ C & MDP (Value iter) & N/A & Implicit & N/A & Learned (tabular)\\ 
            Gordon et al. 2019 \cite{gordonWhatShouldNow2019} & M P & Symbolic (PDDL) & Complete  & Provided & Learned (deep) & N/A \\ 
            Illanes et al. 2020 \cite{illanesSymbolicPlansHighLevel2020} & P M C & Symbolic (PDDL) & Abstract & Provided & Learned (tabular) & Learned (tabular)\\ 
            Leonetti et al. 2016 \cite{leonettiSynthesisAutomatedPlanning2016}& P C & Symbolic (ASP) & Abstract & Provided & N/A & Learned (tablular)\\ 
            Yang et al. 2018 \cite{yangPEORLIntegratingSymbolic2018} & M P C & Symbolic (ASP) & Abstract & Provided & Learned (tabular) & Learned (deep)\\ 
            Lyu et al. 2019 \cite{lyuSDRLInterpretableDataefficient2019} & M P C & Symbolic (ASP) & Abstract & Learned & Learned (tabular) & Learned (deep)\\ 
            \rowstyle{\bfseries}
            Ours & M P C & Symbolic (PDDL) & Myopic+ Abstract & Learned & Learned (deep) & Learned (deep)\\ 
        \end{tabular}
\end{center}
\end{table*}

\subsection{Model-based RL}
The use of model-based planning to increase the sample efficiency of RL dates back to the Dyna architecture \cite{suttonIntegratedArchitecturesLearning1990}.
With the rise of deep RL, most recent model-based work instead uses models in a learned latent space.
These latent models can be applied in a variety of ways, such as: generating experience for a model-free controller \cite{haRecurrentWorldModels2018,hafnerDreamControlLearning2020,hafnerMasteringAtariDiscrete2021}, to supplement a hybrid policy network with a planning-based lookahead \cite{ohValuePredictionNetwork2017,francois-lavetCombinedReinforcementLearning2019,schrittwieserMasteringAtariGo2020}, or as input to a model predictive controller \cite{hafnerLearningLatentDynamics2019}.
While they have achieved impressive results in a range of complex tasks, the need to learn environment models from scratch means they require large amounts of experience to train.
By using an easily specified abstract and myopic domain model we alleviate the need for this long training period.

Another line of relevant recent model-based work uses provided \cite{gopalanPlanningAbstractMarkov2017} or learned \cite{winderPlanningAbstractLearned2020} hierarchical planning models to further increase efficiency and allow transfer of components to related tasks in a tabular environment.
Similarly Roderick et al. \cite{roderickDeepAbstractQnetworks2018} define an abstract symbolic representation which is used to communicate goals to a low-level controller that uses deep RL to learn options.
Unfortunately these methods cannot scale to environments with large abstract state spaces as the meta-controller still uses tabular count-based models.



\subsection{Symbolic planning and RL}
Most prior work integrating \emph{symbolic} planning and reinforcement learning is similar to HRL except the meta-controller has been replaced by a planner. 
Recognising the burden of specifying a complete domain model, there has been a long-lasting focus on using abstract models \cite{ryanUsingAbstractModels2002,groundsCombiningReinforcementLearning2005, illanesSymbolicPlansHighLevel2020}.
The controller learns options for each of the planning actions, and these are chained together at runtime by a plan constructed over the abstract domain.

In this tradition, the most closely related work to ours is SDRL \cite{lyuSDRLInterpretableDataefficient2019}, and its precursor PEORL \cite{yangPEORLIntegratingSymbolic2018}. 
These works extend the use of abstract planning models to the deep RL setting, while also allowing the controller to learn the expected values of executing each option in the environment.
This allows the planner to take into account rewards that are unknown in the original planning model, and avoid actions which are unable to be executed reliably by the controller. 
Like SAGE, they have a three-tier architecture with a meta-controller guiding a symbolic planner, but the meta controller simply sets a numeric reward threshold the plan must exceed, not a symbolic state to aim for.
Furthermore, they assume that each episode terminates after a single plan, and that the agent has a single unique starting state, limiting their applicability to a narrow set of domains.
We elaborate on this comparison in \cref{sec:benchmarking}.

An alternate approach is to use the output from a planner to modify the MDP given to the learning agent. 
DARLING uses a model (without rewards) of the state space to generate a set of plans that are close to optimal in the number of planning steps \cite{leonettiSynthesisAutomatedPlanning2016}.
Any action that does not appear in these plans is occluded from the agent, lowering sample complexity by reducing the search space.
This presupposes that short plans are close to optimal in terms of environmental reward, which may not be the case. 
Recently Illanes et al. \cite{illanesLeveragingSymbolicPlanning2019} combine HRL with symbolic planning, and similarly prune unhelpful options from consideration during execution. 

None of these approaches are directly applicable to our problem, as they require a single planning goal to be specified at the start of the episode.
While this can be mitigated by replanning periodically throughout the episode, they are all inherently reactive. 
The agent is restricted to respond to what is currently visible; there is no mechanism for the planner to predict and prepare for future events that have not occurred. 
By contrast, our technique allows online ancillary goal generation during episodes to allow the agent to anticipate events which are not captured by its myopic model. 

Gordon et al. \cite{gordonWhatShouldNow2019} propose an architecture, HIP-RL, which has some similarities to our approach; an RL-based meta-controller can call a symbolic planner to produce plans to achieve its goals.
A critical point of distinction is that in their work, the goal for the planner is fixed and pre-determined by a human at the start of every episode.
This limits the flexibility of the system; all the meta-controller can do is choose to plan towards the pre-defined goal or not, whereas SAGE can generate its own goals for the planner according to the needs of the situation.
See \cref{tab:relatedwork} for a concise overview of the hybrid planning and learning systems discussed above.


\section{SAGE}

\begin{figure}
    \centering
    \begin{tikzpicture}[node distance=1.8cm and 2cm]
        
        \node [zone, minimum width=7.6cm, minimum height=2.8cm, fill=black!40] at (1.25,1.05) (inner) {};
        \node [zone, minimum width=7.8cm, minimum height=4.6cm] at (1.25,1.9) (medium) {};
        \node [zone, minimum width=8.0cm, minimum height=8.2cm] at (1.25,3.65) (outer) {};
        \node[label] at ([shift={(0.8,0)}]inner.north west) {\texttt{\textbf{Inner}}};
        \node[label] at ([shift={(0.8,0)}]medium.north west) {\texttt{\textbf{Middle}}};
        \node[label] at ([shift={(0.8,0)}]outer.north west) {\texttt{\textbf{Outer}}};

        \node (environment) [environment2] {Environment};
        \begin{scope}[node distance=1.9cm]
            \coordinate[right of=environment] (d1);
        \end{scope}
        \begin{scope}[node distance=2.2cm]
            \coordinate[left of=environment] (d2);
        \end{scope}

        \node (bottom) [module, above of=d1] {Low-level controller};
        \node (queue) [action, above of=bottom] {Queue};

        \foreach \x/\y in {-0.1/0.1, 0.1/-0.1}{
        \node[action] at ([shift={(\x,\y)}]queue) {};
        }

        \node (middle) [module, above of=queue] {Symbolic planner};
        \node (top) [module, above of=middle] {Meta-controller};

        \coordinate[above of=d2] (d3);
        \coordinate[above of=d3] (d4);
        \coordinate[above of=d4] (d5);
        \coordinate[above of=d5] (d6);
    
        \draw [arrow] (top) --  node[anchor=east] {ancillary goal} node[anchor=west,text width=3cm] {``deliver passenger"} (middle);
        \draw [rarrow] (queue.north)+(0,0.1) -- node[anchor=east] {plan }  node[anchor=west,text width=3cm] {(symbolic steps)} (middle);
        \draw [arrow] (queue.south)+(0,0.1) --  node[anchor=east] {next step in queue}node[anchor=west,text width=3cm] {``take second right"} (bottom);

        \draw [arrow] (bottom.south) node[anchor=north east] {low-level actions} node[anchor = north west, text width=3cm] {``brake and indicate"}  |- (environment.east);
        \draw [line] (environment.west) -| (d6);

        \draw [arrow] (d3) node[anchor = north west,shift={(0,-1)}] {observations}
         -- (bottom.west);

        \draw [arrow] (d5) -- (middle.west);    
        \draw [arrow] (d6) -- (top.west);   
    
    \end{tikzpicture}
    \caption{SAGE conceptual architecture showing data flow. The three background boxes correspond to the loops in \Cref{alg:nested}.}
\label{fig:architecture}
\end{figure}
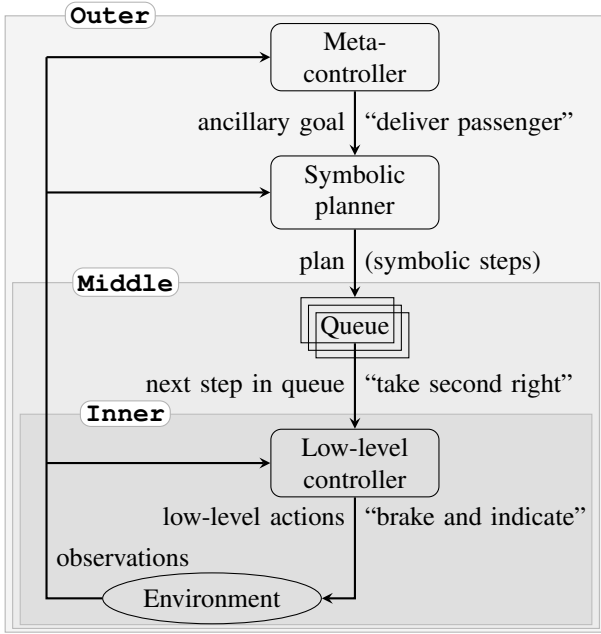

SAGE has a three tier hierarchical architecture (\cref{fig:architecture}):
\begin{enumerate}
    \item The meta-controller - an RL component which takes state observations and outputs ancillary goals.
    \item The symbolic planner - an off-the-shelf planner which takes ancillary goals and outputs a plan composed of symbolic steps.
    \item The low-level controller - an RL component which takes an individual symbolic step and state observation, and outputs atomic actions in the environment.
\end{enumerate}

To understand how these components interact, we return to the taxi example.
The meta-controller sees the current map, and needs to make a decision about what to do next.
In this scenario, its options are to pick up a new passenger, move to a particular location, deliver a passenger already in the taxi, or some combination of these. 
The chosen ancillary goal is then passed to the planner, which uses a symbolic representation of the map to construct an abstract symbolic plan to achieve that goal. 
This plan consists of high-level steps such as ``head to the end of the road, pick up the passenger there, then make two right turns."
Each step of the plan is passed in sequence to the low-level controller, which steers and accelerates the car to achieve the current step.
Both the low-level and meta-controllers are trained by interacting with the environment. 

\begin{figure}
    \includegraphics{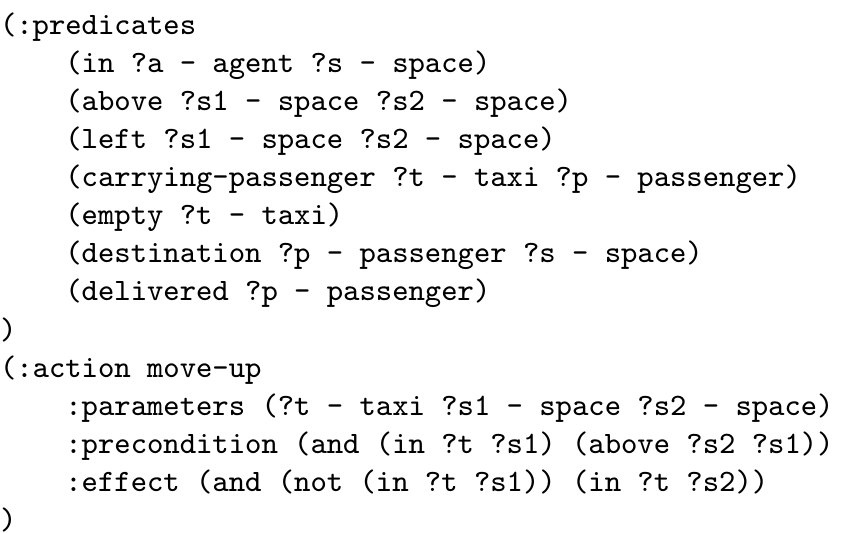}      
\caption{Excerpt of the PDDL domain file for the taxi domain.
    See code repository for full files: \url{https://github.com/AndrewPaulChester/sage-code/blob/main/planning/taxi/domain.pddl}.}
\label{fig:pddl}
\end{figure}

In line with prior work \cite{lyuSDRLInterpretableDataefficient2019} we assume access to a symbolic planning domain model (\cref{fig:pddl}) and a high-level (PDDL) state representation. 
Unlike in most prior work, this is \emph{not an onerous burden} as our approach allows the use of myopic and abstract models.
This allows the designer to omit dynamics from the model which are unknown or difficult to capture formally, and focus solely on the aspects of the problem which are amenable to symbolic reasoning.
In the remainder of this section we formalise each component before presenting the unified planning and learning algorithm.

\subsection{Meta-controller}
The meta-controller is responsible for the strategic direction of the agent.
It acts on an abstract semi-MDP with modified action space, transition and reward functions.
Since the actions correspond to goals in the symbolic planning language, for each predicate in the planning domain we include a binary output which determines if it is included in the goal, as well as a variable number of discrete or continuous outputs which determine the arguments for that predicate (see \cref{sec:experiments} for examples). 

The meta-controller acts on an extended time scale; it receives a single transition when the entire symbolic plan has been completed (or abandoned due to time out).
That is, an experience step is of the form $(s_t, G_t, U_{t:t+n},s_{t+n})$, where $s_t,s_{t+n}$ are the state observations at the start and end of the plan respectively, $G_t$ is the action (i.e. ancillary goal) chosen, and $U_{t:t+n} = \sum_t \gamma^{t} r_{t}$ is the accumulated environmental reward over the plan of duration $n$.
This results in a slow rate of experience generation, we outline some techniques to accelerate learning in \cref{sec:learning-speed}.
SAGE is not restricted to a single learning algorithm, and we show results for both on- and off-policy approaches; for generality we assume experience is stored in a replay buffer, $\mathcal{B}$ to train on.

\subsection{Symbolic planner}
The planner decomposes the ancillary goals ($G$) set by the meta-controller into a series of achievable symbolic steps ($o\in \ActsP$).
The planning domain is fixed and provided up front, so each new plan only requires a description of the initial and goal states.  
The planning goal is determined by the meta-controller as described above, while the initial state is the high-level PDDL description of the current state observation.
Given the planning instance, the symbolic planner outputs a sequence of symbolic steps required to achieve the goal. 
These steps are stored in a queue, and each is passed one by one to the low-level controller for use in goal-directed RL.
Once the final step in the queue has been completed, we deem the plan to be successful; otherwise if any individual step of the plan timed out, we say the plan failed.

\subsection{Low-level controller}
The low-level controller operates on a universal MDP \cite{schaulUniversalValueFunction2015} $\MDP_L = (\States,\ActsP,\Acts,\Trans,\Rewards_L)$, where $\ActsP$ is the set of possible goals which correspond to the symbolic steps given by the planning model.
The reward function $\Rewards_L : (\States,\ActsP,\Acts) \rightarrow \RR$ is equal to the environment reward plus an intrinsic bonus of $r_{complete}$ when the step has been completed.
This encourages the controller to achieve the intended goal while still allowing it to be sensitive to the underlying reward signal.
The state space $\States$, action space $\Acts$, and transition function $\Trans$ are unchanged from the base environment. 

The low-level controller must be able to determine when it has successfully completed the symbolic step so it can request the next one from the queue. 
This is done by checking whether the postconditions of the current step are met in the high-level PDDL state description of the current observation.
In addition, in order to prevent the agent from getting stuck trying to complete a step which is beyond the controller's capabilities, we implement a timeout limit of $\tau_{max}$ atomic actions per symbolic step. 
If the step has not been completed by then, the entire plan fails and control passes back to the meta-controller to choose a new goal. 

We use PPO as our learning algorithm for the low-level controller \cite{schulmanProximalPolicyOptimization2017}, and we stress that this is independent of the choice of algorithm for the meta-controller, as they use different underlying experience to train on.
In our experiments we have pre-trained the low-level controller to improve the convergence of the meta-controller.

\begin{figure}[!t]
    \begin{algorithm}[H]
    \caption{Conceptual sketch of main control loop }
    \label{alg:nested}
    \begin{algorithmic}[1]
        \For {$k = 1,...,total\_episodes$}
            \While {episode not finished} \hfill \texttt{\textbf{Outer}}
                \State $G \sim \pi_G(\cdot | s ; \theta_G)$ \Comment{Generate ancillary goal} \label{timeout}
                \State  $P \gets \texttt{plan}(s,G)$ \Comment{Create symbolic plan}
                \For {\textbf{each} step $o$ in $P$} \hfill \texttt{\textbf{Middle}}
                    \State $\tau \gets \tau_{max}$ \Comment{Initialise step timeout}
                    \While{step $o$ not completed} \hfill \texttt{\textbf{Inner}}
                        \State \textbf{if} $\tau = 0$ \textbf{go to} 3
                        \State $a \sim \pi_L(\cdot | s, o; \theta_L)$ \Comment{Get atomic action}
                        \State $s,r \gets \texttt{act}(a)$ \Comment{Act in environment}
                        \State $\theta_G \gets \texttt{train}(\mathcal{B})$ \Comment{Perform training}
                        \State $\tau \gets \tau-1$
                    \EndWhile
                \EndFor
            \EndWhile
        \EndFor
    \end{algorithmic}
    \end{algorithm}
\end{figure}

\subsection{Learning algorithm}
For each episode, the learning algorithm takes the form of three nested loops. 
At the innermost level, the low-level controller acts to complete the current symbolic step.
In the middle level, the planner provides a new symbolic step when the previous one is complete.
At the outermost level, the meta-controller chooses a new ancillary goal after the entire plan is finished.
The pseudocode is presented in \Cref{alg:nested}.

\subsection{Improving learning speed}
\label{sec:learning-speed}
The meta-controller gathers experience more slowly than the low-level controller, especially in environments with many atomic actions per ancillary goal.
This can result in very slow progress, despite a relatively high sample efficiency, as the clock time per frame of experience is high. 
This section outlines some additional steps taken to improve the speed of training for the meta-controller.

Firstly, we have augmented the experience of the meta-controller with interim states in each plan.
Consider a taxi that is driving to the airport; the experience gained not only applies to trips from the starting point to the airport, but also to all trips to the airport from any of the interim positions of the taxi. 
Formally, in addition to the standard experience step for a plan of length $n$: $(s_{t}, G_t, U_{t:t+n},s_{t+n})$, we also store steps which correspond to the observations from the middle of the plan, i.e. $(s_{t+i}, G_t, U_{t+i:t+n},s_{t+n})$ for $0<i<n$.

\begin{figure}
    \centering
    \begin{tikzpicture}[font=\footnotesize, node distance=1cm]
        
        \node (o1) [observation] {};
        \node (o2) [observation, right of=o1] {};
        \draw [arrow] (o1) node[anchor=north] {$s_t$} -- node[anchor=south] {$a_t$} (o2);   
        \node (o3) [observation, right of=o2] {};
        \draw [arrow] (o2) node[anchor=north] {$s_{t+1}$} -- node[anchor=south] {$a_{t+1}$}  (o3);   
        \node (o4) [observation, right of=o3] {};
        \draw [arrow] (o3) node[anchor=north] {$s_{t+2}$}-- (o4);   
         
        \node (blank) [right of=o4]{\ldots};
        \draw [arrow] (o4) node[anchor=north] {$s_{t+3}$} -- (blank);  

        \node (o5) [observation, right of=blank] {};
        \draw [arrow] (blank) -- (o5) node[anchor=north] {$s_{t+n-2}$};   
        \node (o6) [observation, right of=o5] {};
        \draw [arrow] (o5) -- (o6);   
        \node (o7) [observation, right of=o6] {};
        \draw [arrow] (o6) --  (o7) node[anchor=north] {$s_{t+n}$};   

        \draw [arrow] (o1) to [bend left=70] node[anchor=south] {$G_t,U_{t:t+n}$} (o7);   
        \draw [arrow,dashed] (o3) to [bend left=50] node[anchor=south east] {$G_t,U_{t+2:t+n}$} (o7);  
        \draw [arrow,dashed] (o5) to [bend left=30]  (o7);   
 
    \end{tikzpicture}
\caption{Visualisation of interim experience with interval $k=2$. The solid curved line is the original experience step $(s_{t}, G_t, U_{t:t+n},s_{t+n})$, the dashed lines are interim experience steps.}
\label{fig:interim}
\end{figure}
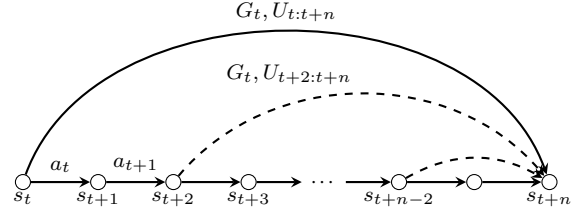

We found that taking interim experience from every intermediate frame yielded too many similar samples and resulted in unstable learning.
Instead we took experience only at every $k^{th}$ frame, shown in \cref{fig:interim} with $k=2$.
This interim experience is off-policy, which restricts our choice of learning algorithms for the meta-controller. 

Secondly, we can exploit the knowledge of the symbolic planner to provide feedback for invalid goals.
That is, if the meta-controller sets a goal which is not possible to achieve according to the planning model, we can identify this immediately and penalise the meta-controller.
This feedback is combined with the rewards received from the environment, as seen in the following equation: 
\begin{equation} 
    U_{t:t+n} = \begin{cases}
    \sum\limits_{t'=t}^{t+n} \gamma^{t'} r_{t'} & \text{if plan found}\\
        -r_{invalid} & \text{if no valid plan found}\\
    \end{cases}
\end{equation}  
This is a large advantage over non-symbolic hierarchical methods which must wait until the low-level controller tries and fails to achieve an impossible goal.
If no valid plan was found, the agent executes a single no-op action in the environment and generates a new ancillary goal next step.

\section{Experiments}
\label{sec:experiments}
SAGE allows the use of abstract, myopic models, which no other method can use directly.
Consequently, we cannot simply benchmark against a prior method and provide a direct comparison with identical inputs.
Instead, in this section we wish to demonstrate that prior methods are unable to effectively learn in our environments, and that SAGE, by exploiting the easily defined myopic models, can.
Specifically, the hypotheses we test with our experiments are:
\begin{enumerate}[label=H\arabic*. ,wide=0pt]
	\item SAGE can achieve higher scores than a scripted planning approach by learning regularities in the environment which are not part of the provided model.
	\item SAGE can scale to larger domains than other learning approaches by exploiting its provided myopic model.
    \item SAGE requires the planner to learn effectively, not just the hierarchical decomposition of the problem.
\end{enumerate}
By providing evidence for these hypotheses, we show that SAGE can effectively solve environments that are out of reach of prior approaches.

Our first experimental domain, an extension of the classic taxi gridworld setting \cite{dietterichHierarchicalReinforcementLearning2000}, uses a myopic world model that is not abstract. 
This allows us to focus on ancillary goal generation without the complications of low-level control, and addresses hypotheses 1 and 2. 
Second is a 2D Minecraft inspired world similar to those used in other work \cite{andreasModularMultitaskReinforcement2017,illanesSymbolicPlansHighLevel2020}.
We have extended the environment from a grid world to a simple physics based motion system and provided a model which is both abstract and myopic. This showcases the full three tier architecture of SAGE, addressing hypotheses 2 and 3.
We begin this section with a detailed discussion of the methods we have used for comparison.
We then introduce the dynamics of each domain and outline the provided myopic model before presenting and discussing the empirical results.
Hardware details and hyperparameters are provided in \cref{sec:reproducibility}.

\subsection{Benchmarking}
\label{sec:benchmarking}
We compare SAGE's performance against five other methods, each carefully chosen to provide the most comparable benchmarks.
See \cref{tab:benchmarks} for a high level summary.

\subsubsection{Scripted}
The scripted method is a planning approach which uses the same planning model as SAGE.
It requires a hand-chosen planning goal, which is to always deliver the passenger to their destination (if there is a passenger present).
This tests hypothesis 1, and is only applicable to the Taxi domain as there is no scripted low-level controller in the CraftWorld domain.

\subsubsection{PPO}
Proximal Policy Optimisation (PPO) is a standard model-free RL algorithm which uses the same network architecture as our meta-controller.
It does not have access to the planning model, and serves as the state-of-the-art `vanilla' DRL method, showing that these problems are too challenging for standard techniques.

\subsubsection{VPN}
Value Prediction Network (VPN) \cite{ohValuePredictionNetwork2017} is a strong model-based RL method, which is an open-source precursor to MuZero \cite{schrittwieserMasteringAtariGo2020}.
It does not have access to the planning model, but has the capacity to learn its own model through experience. 
This demonstrates that the provided model, while easy for humans to specify, cannot be learned efficiently by current methods. 

\subsubsection{SDRL}
Symbolic Deep Reinforcement Learning (SDRL) \cite{lyuSDRLInterpretableDataefficient2019} is a symbolic planning-RL hybrid system, and is the most closely related prior work.
SDRL as specified in the original paper is inapplicable to our problem as it makes two strong assumptions that do not hold in our domains: 
\begin{enumerate}
  \item there is a singular starting state for each domain
  \item the episode terminates after the completion of a single plan.
\end{enumerate}
These assumptions are implicit in the form of the algorithm; SDRL stores the latest plan found by the planner ($\Pi$) to use as a backup. 
If the planner cannot meet the new quality constraint, then it reverts to executing the previous plan.
This is only possible if every planning step starts in the same state.
If the start state is random, or plans must be constructed in the middle of an episode, one cannot reuse the last plan that was executed, as the actions may be unsuitable for the current state.
In order to apply SDRL to our domains where these assumptions do not hold, we must now store $\Pi(s)$, that is a different plan for every symbolic state.
This is one of the ways in which the SDRL meta-controller is inherently tabular in the symbolic state space. 
Accordingly, SDRL cannot scale to our domains except for small taxi, which has only 525 symbolic states.
Our other domains have many ($>10^{12}$) symbolic states, as the random road network is encoded symbolically in the taxi domain, and the symbolic CraftWorld domain tracks the quantities of various resources.

\subsubsection{hDQN}
Hierarchical Deep Q-Network (hDQN) \cite{kulkarniHierarchicalDeepReinforcement2016} is a two layer hierarchical RL method.
To make the comparison as precise as possible, hDQN's controller receives the same pre-trained skills as SAGE.
As a result, this is only applicable in the CraftWorld domain, because the taxi domain does not have a low-level controller. 
This method is equivalent to an ablated version of our full architecture without the planning layer, which tests hypothesis 3. 
We have specifically chosen this method as, like SAGE, it can incorporate pre-trained skills in its controller, and admits a manual decomposition of the task into the two layers.
This makes it a more challenging benchmark and a fairer comparison.



\begin{table}
    \caption{Summmary of Benchmarks.\\ *SDRL only applicable to small taxi domain.}
    \begin{center}
        \begin{tabular}{c |c c c c}
        & Taxi & CraftWorld & Model provided & Hypothesis\\\hline
        Scripted & \checkmark & & \checkmark & H1\\
        PPO & \checkmark & \checkmark & & H2\\
        VPN & \checkmark & \checkmark & & H2\\
        SDRL & \checkmark* & & \checkmark & H2\\
        hDQN & & \checkmark & & H3\\
        \end{tabular}
    \end{center}
    \label{tab:benchmarks}
\end{table}

\subsection{Taxi domain}

\begin{figure}[!t]
    \centering
    \includegraphics[height=2.3in]{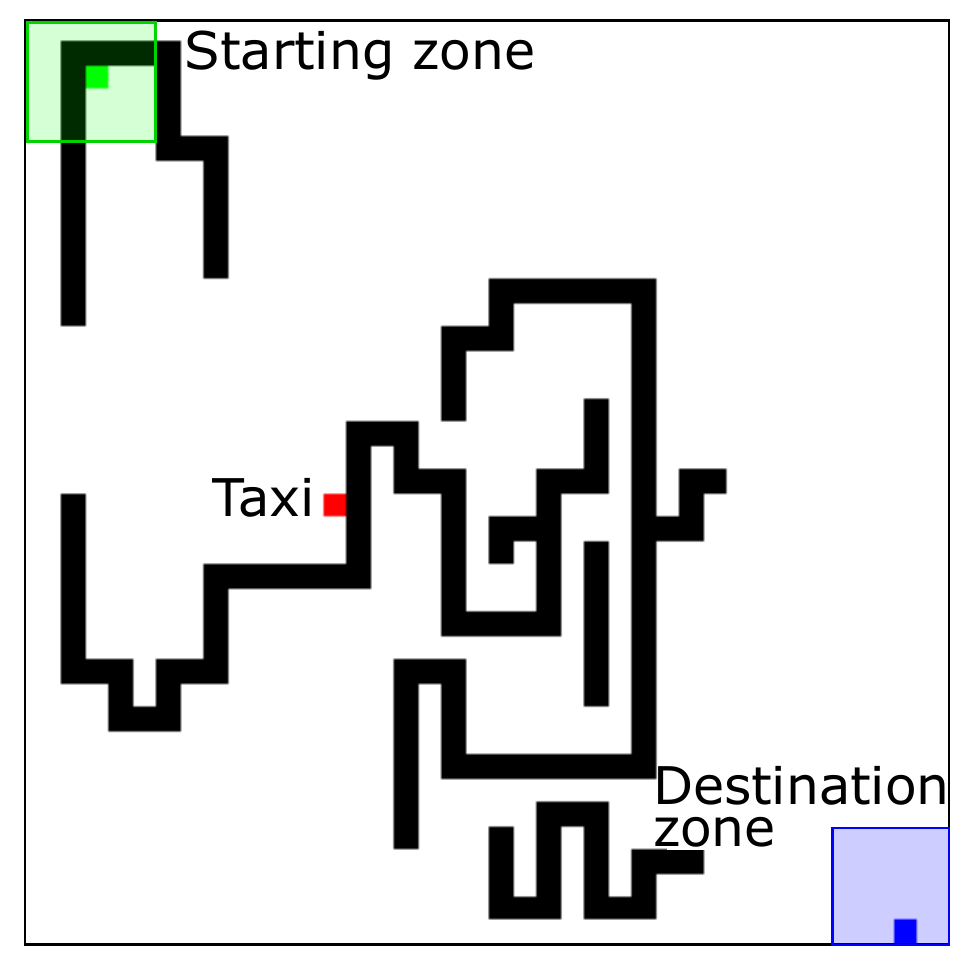}%
    \caption{Large taxi domain (20x20). The black walls are randomly generated at the start of every episode.}
    \label{fig:taxi-env}
\end{figure}

\begin{figure*}[!t]
    \centering
    \hfil
    \subfloat[]{\includegraphics[height=1.5in]{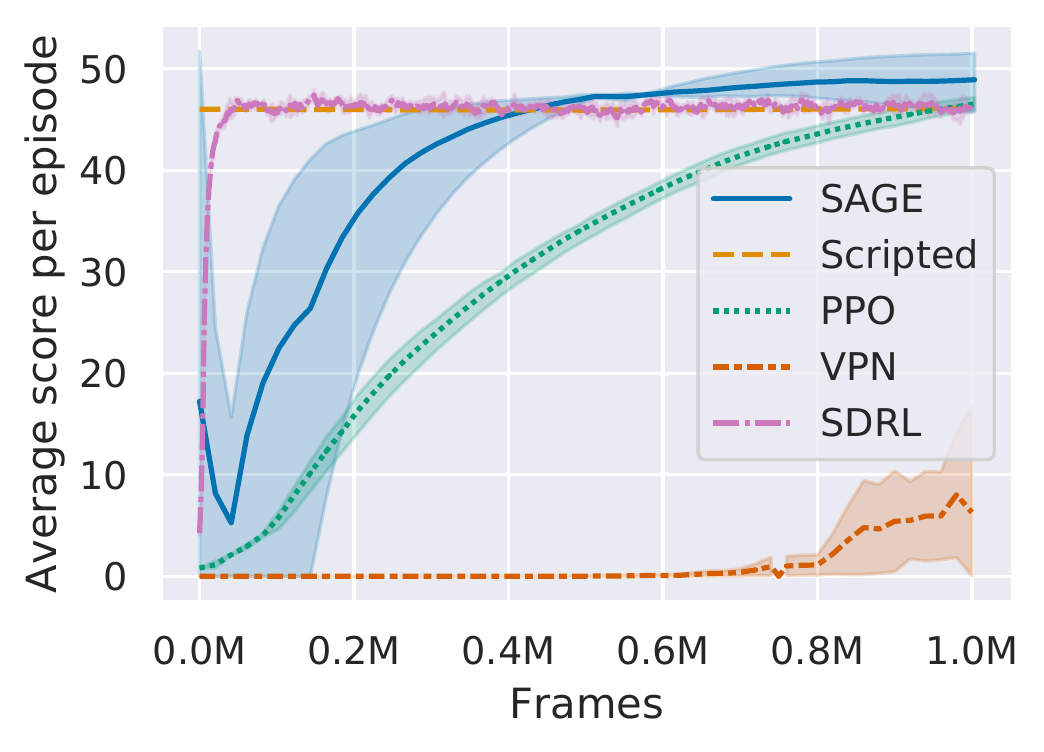}%
    \label{fig:small-taxi}}
    \hfil
    \subfloat[]{\includegraphics[height=1.5in]{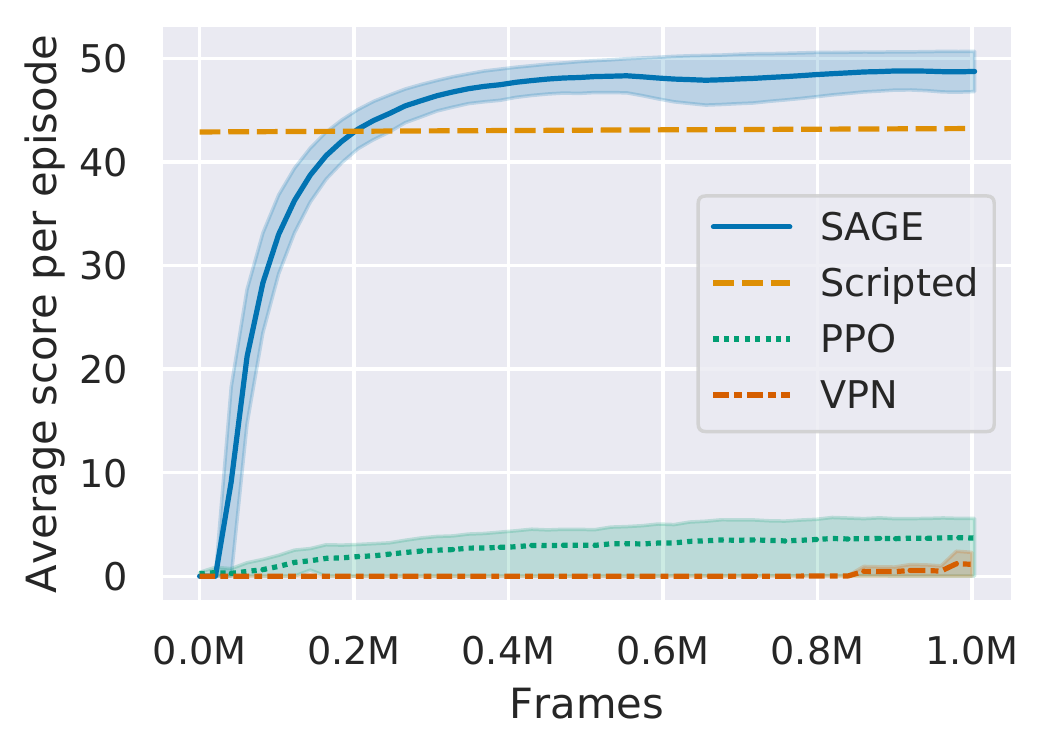}%
    \label{fig:medium-taxi}}
    \hfil
    \subfloat[]{\includegraphics[height=1.5in]{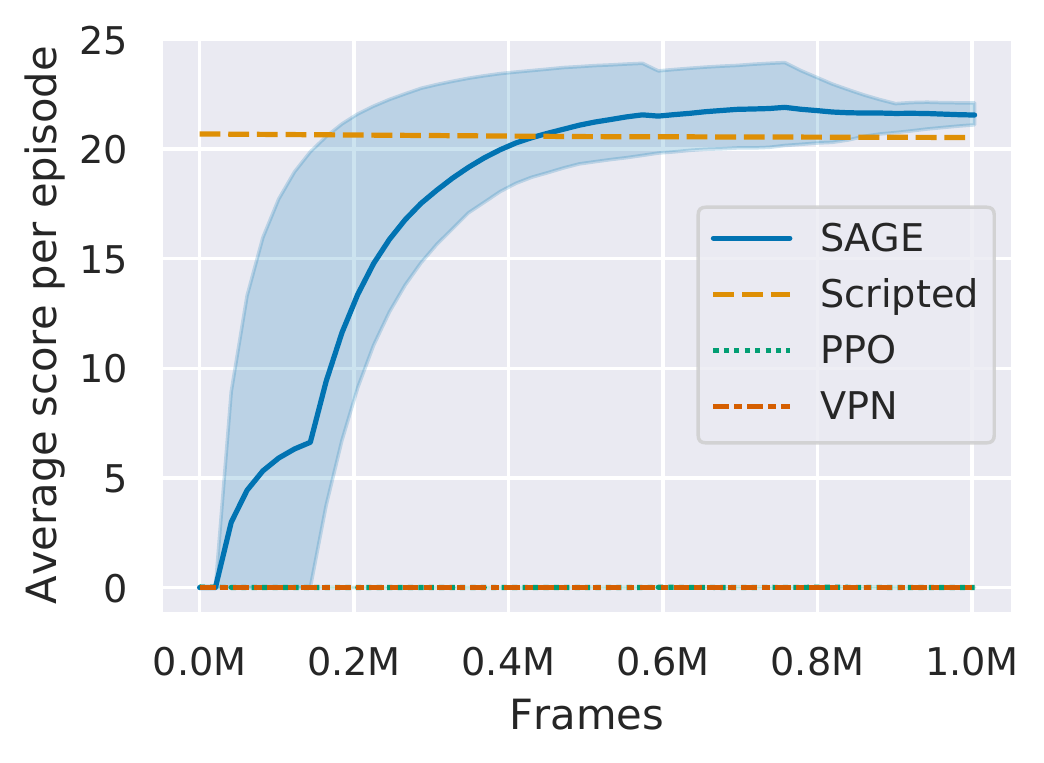}%
    \label{fig:large-taxi}}
    
    \hfil
    \subfloat[]{\includegraphics[height=1.5in]{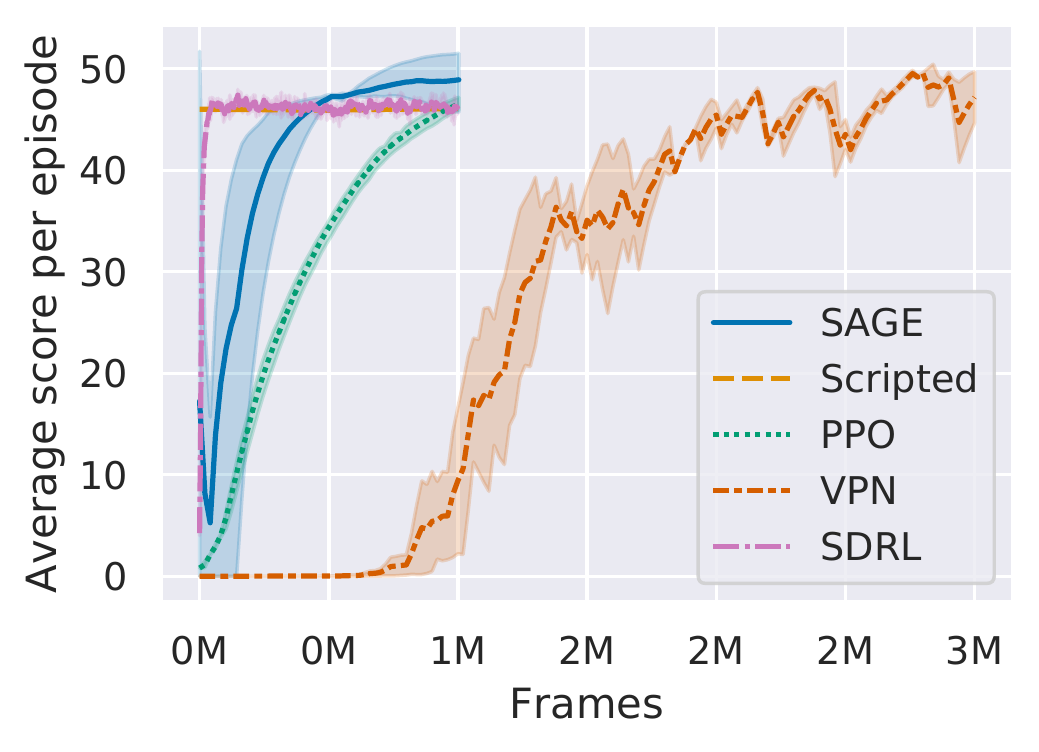}%
    \label{fig:small-taxi-sup}}
    \hfil
    \subfloat[]{\includegraphics[height=1.5in]{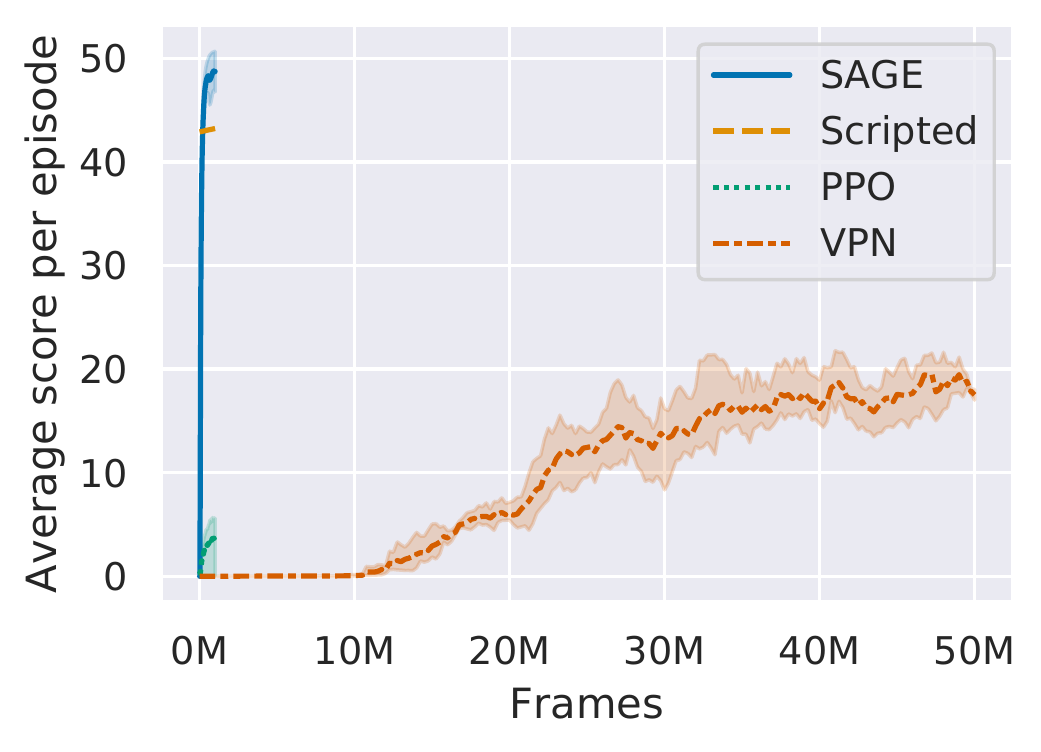}%
    \label{fig:medium-taxi-sup}}
    \hfil
    \subfloat[]{\includegraphics[height=1.5in]{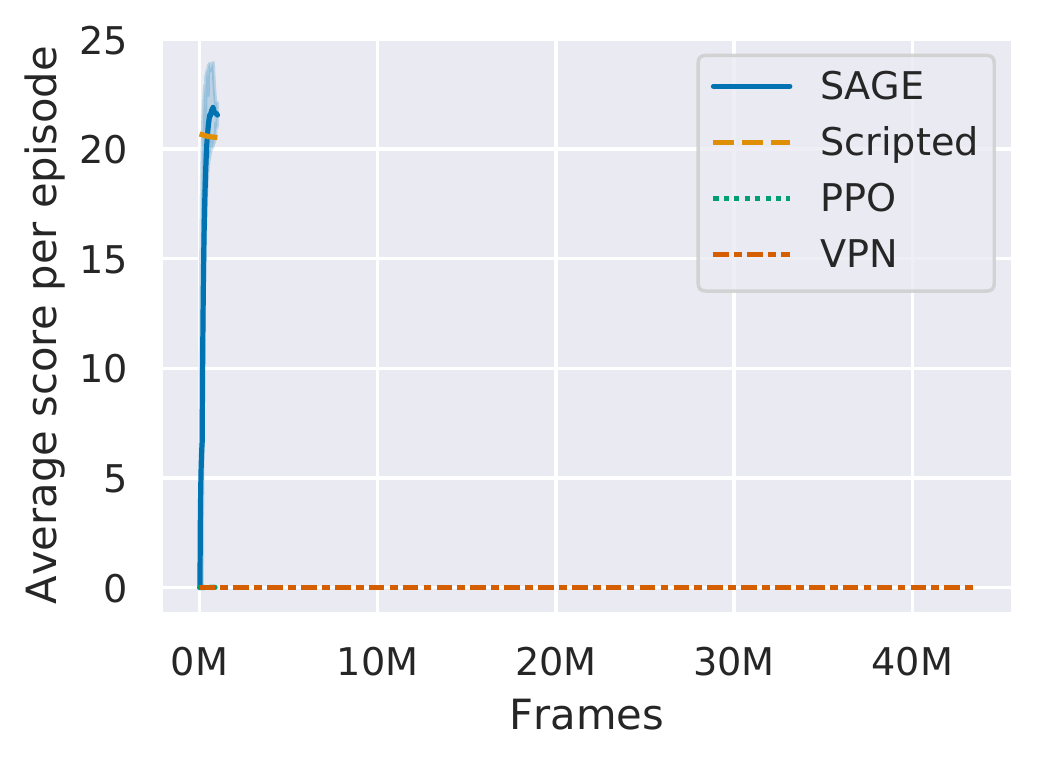}%
    \label{fig:large-taxi-sup}}
    \caption{Results achieved in Taxi domain in the small (a), medium (b) and large (c) environments. The bottom row (d-f) shows extended results demonstrating VPN learns on small and medium environments after longer time periods}
    \label{fig:results}
    \end{figure*}

A $n\times n$ grid world (\cref{fig:taxi-env}) is randomly populated with a sparse maze at the start of each episode, and the taxi starts in a random unoccupied space. 
At each timestep if there is no passenger on the map or in the taxi, another passenger is spawned with probability $p$.
The passenger's location is sampled uniformly from the starting zone, a $k\times k$ square at the top left of the map, with their destination sampled from a corresponding destination zone in the bottom right. 
A reward of 1 is given for each passenger delivered to their destination, and each episode is terminated after $t$ timesteps.
To showcase the scalability of our method we perform experiments on small, medium and large versions of this environment, with values $n=(5,10,20)$, $p=(0.12,0.08,0.05)$, $k=(2,2,3)$, $t=(1000,2000,2000)$ respectively.
Due to the limited size of the small environment we do not perform random maze generation, so the position of walls is constant across all episodes in line with the original domain.

The provided myopic model (\cref{fig:pddl}) captures the road network and the requirements for picking up and dropping off passengers, but crucially does not know the location of the starting or destination zones.
The meta-controller acts in a factored hybrid action space with Boolean variables indicating the presence of PDDL predicates, and discrete and continuous variables indicating the corresponding predicate arguments.
For example, the goal \texttt{empty $\wedge$ in(4,6) $\wedge$ delivered(p0)} indicates that passenger 0 should be delivered to their destination, the taxi should be empty and in square $(4,6)$. 
To accommodate this, we have used PPO \cite{schulmanProximalPolicyOptimization2017} as the algorithm for the meta-controller in these experiments as it can handle hybrid discrete and continuous action spaces \cite{fanHybridActorCriticReinforcement2019}. 
We used the Fast Downward planner \cite{helmertFastDownwardPlanning2006} as there were no numeric functions.

\subsection{Taxi results}

As can be seen in \cref{fig:small-taxi,fig:medium-taxi,fig:large-taxi}, SAGE quickly learns a near-optimal policy across all environment sizes. 
This is unsurprising as the ancillary goals the meta-controller needs to learn are equivalent in each case: deliver any existing passenger and then return to the top left to wait for the next one.
The visual input is more complex and varied in larger environments, which can be seen by a slightly longer training time and some minor instability in the large policy. 

As expected, the scripted planner is also unaffected by the environment size, and exhibits strong but sub-optimal performance in all cases. 
Since it has no mechanism to learn the regularities in the passengers' starting locations it waits in the bottom right of the map after delivering each passenger, rather than returning to the starting zone. 

For the other methods though, scaling the environment up introduces substantial challenges. 
SDRL learns very quickly, but cannot reliably outperform the scripted agent. 
While it can learn how good certain actions are, the planning component is still inherently reactive and cannot foresee future passengers.
The learning speed can be partially attributed to a different observation space; the meta controller requires tabular input, as opposed to the image based input of the other methods. 
This helps it to learn rapidly in small environments, but prevents it from being applied to the larger domains.

PPO learns relatively quickly in the small environment, as random policies often produce rewards, whereas in the large environment it fails to ever get sufficient rewards under the random policy to kickstart learning. 
The medium environment is an interesting middle ground; the agent learns to pick up passengers from the top left and deliver them to the bottom right, but cannot reliably navigate around the randomised walls and so gets permanently stuck in many episodes.

Similarly, while VPN can eventually learn reasonable policies in the small and medium environments (\cref{fig:small-taxi-sup,fig:medium-taxi-sup}), it requires far more experience to learn the model. 
Furthermore, the planning horizon of VPN was set at 10 for these experiments (increased from a maximum of 5 in the original paper), and as the large environment requires plans of up to 80 steps before seeing a reward, even with an accurate model it is challenging for VPN to operate effectively (\cref{fig:large-taxi-sup}).

\subsection{CraftWorld domain}
\label{sec:craftworld-domain}

\begin{figure}[!t]
    \centering
    \includegraphics[height=2.3in]{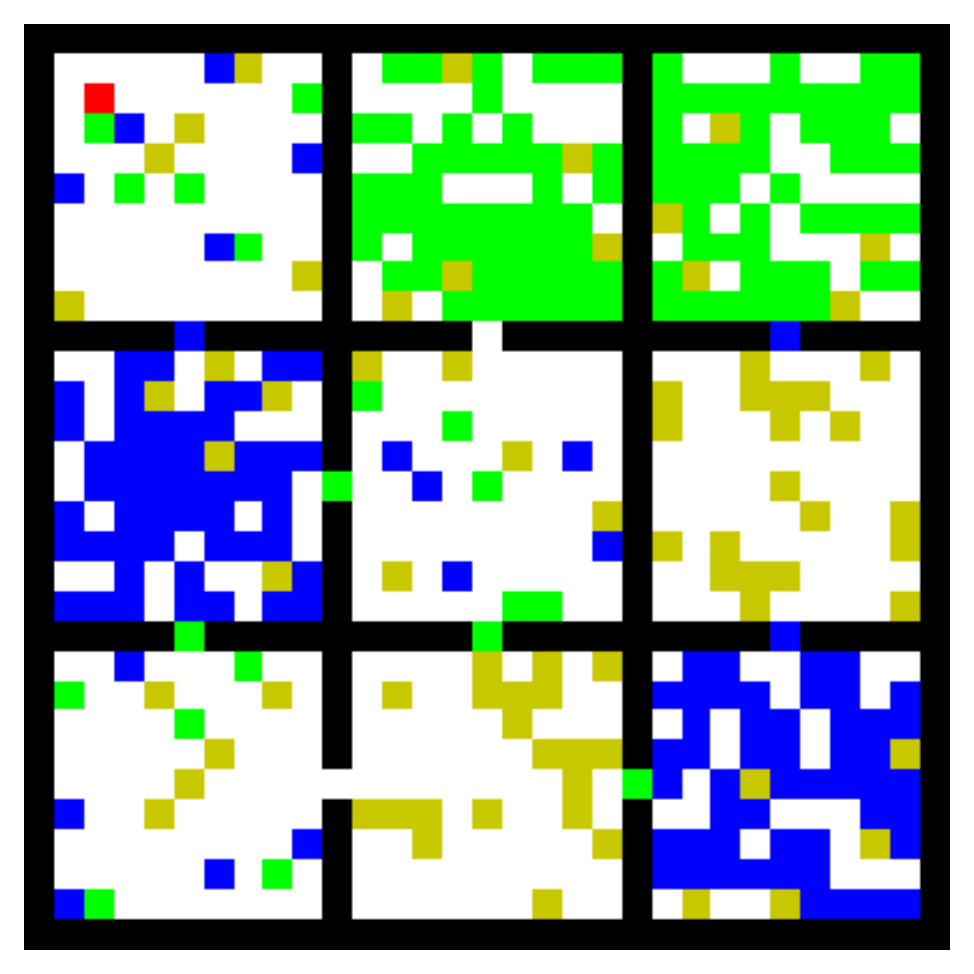}%
    \caption{CraftWorld domain. The agent is red, coins are yellow, trees are green, and stone is blue. Black cells are unbreakable walls. Despite the visual similarity to the taxi domain, in Craftworld the agent turns and accelerates in a continuous space, instead of grid based movement.}
    \label{fig:craft-env}
    \end{figure}

CraftWorld is a 2D domain where the agent is trying to navigate through its surroundings to collect coins (\cref{fig:craft-env}).
The randomly generated environment is filled with a variety of blocks of different materials, some of which can be cleared with the right equipment crafted from other blocks.
For example, the agent can clear tree blocks to collect wood to make a pickaxe, which will then allow it to remove a stone block.

Each level consists of a set of $3\times 3$ rooms, which are separated by walls with a single gap (the door). The door may be empty, or it may be blocked by any of the three different obstacles. 
There are four different types of rooms, each of which has a different proportion of obstacles and coins inside them.
The type of each room and the location of obstacles within each room are randomised at the start of every episode.
Collecting a coin provides a reward of 1, and all other actions provide 0 reward.

The key difference between the CraftWorld and taxi domains is that the agent moves continuously rather than in discrete steps.
Like in Minecraft, the agent turns left and right and accelerates to move in the direction it is facing. 
This is designed to mimic many real world tasks in that the low-level dynamics are unable to be captured accurately by a symbolic model, i.e. the model is abstract, as well as myopic.

The myopic, abstract model we provide for the CraftWorld domain is limited to the information that is easy to describe symbolically, consisting of:
\begin{itemize}
    \item The layout of the rooms, including the obstacle (if any) in each door
    \item The equipment requirements to clear obstacles (e.g. a wooden pickaxe is needed to remove rocks)
    \item The recipes for creating equipment (e.g. 3 planks and 2 sticks are required to create a wooden pickaxe)
    \item The numerical quantity of each object in each room.
\end{itemize}
In contrast to the taxi domain where the planner uses a detailed map, here it only knows about the macro-level room map. 
Since this model is abstract, we pre-train a goal-directed policy using PPO to act as the low-level controller.
For example policies are trained to ``collect coins" or ``craft a pickaxe."
Because the planner needs to reason about quantities of materials, we use ENHSP \cite{scalaHeuristicsNumericPlanning2016}.

\subsection{CraftWorld results}

\begin{figure*}[!t]
    \centering
    \hfil
    \subfloat[]{\includegraphics[height=2.3in]{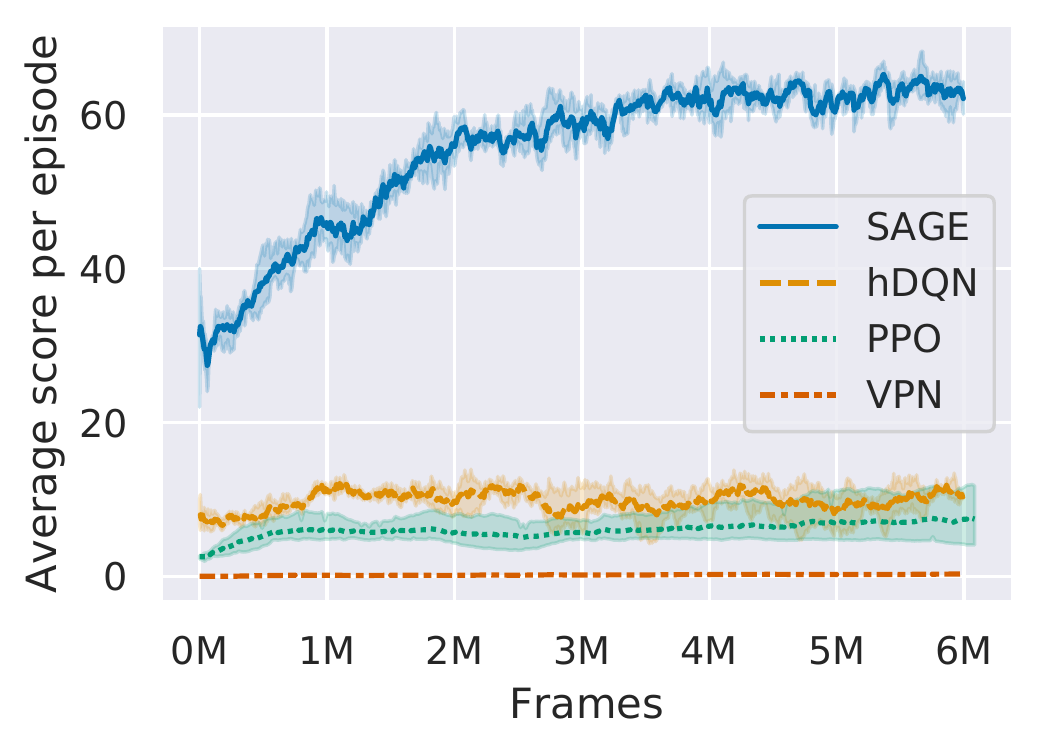}%
    \label{fig:craft}}
    \hfil
    \subfloat[]{\includegraphics[height=2.3in]{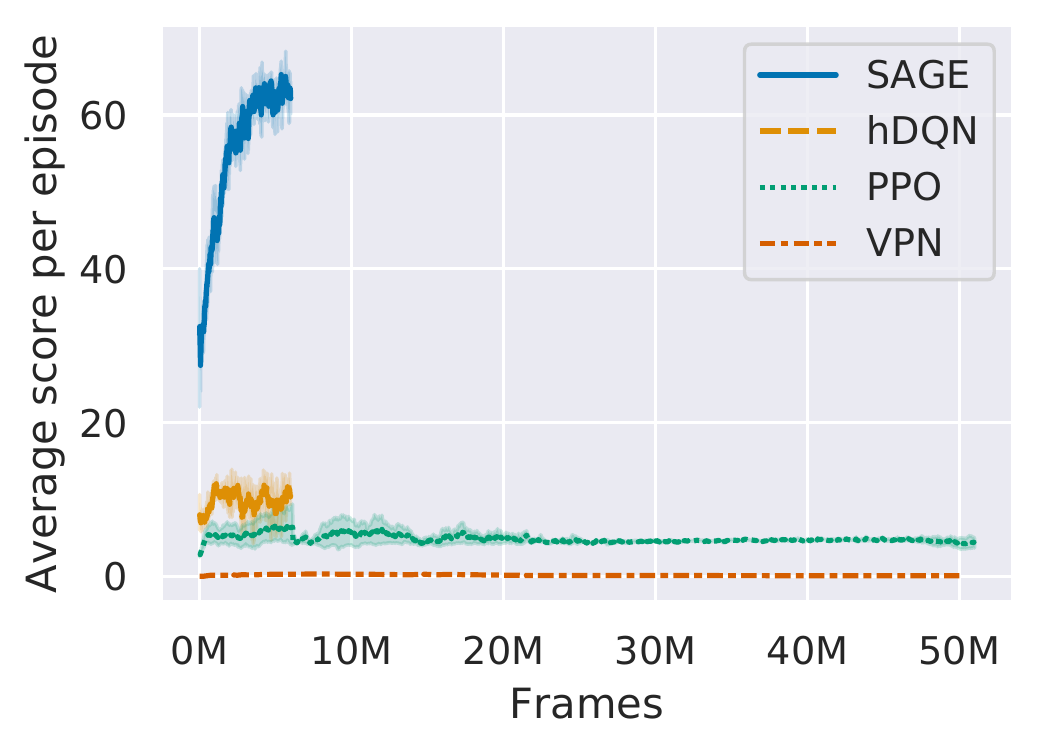}
    \label{fig:extended-craft}}
    \caption{(a) Indicative CraftWorld results for 6M frames of experience. (b) Extended CraftWorld results showing 50M frames of experience for VPN and PPO benchmarks. SAGE and hDQN are only trained to 6M frames as this is on top of 40M frames of pretraining for the low-level controller.}
\end{figure*}

SAGE significantly outperforms the benchmark methods, learning to collect approximately 65 of the 75 total coins in the domain after 6M frames of experience (\cref{fig:craft}).
Neither PPO nor hDQN meaningfully improve after a short initial period; the agents can collect the coins in the first room but frequently get stuck there and cannot reliably navigate between rooms. 
It should be noted that hDQN and SAGE have the benefit of 40M frames of pretrained low-level controllers, so a direct comparison to PPO and VPN in this graph is not possible. 
We ran these methods up to 50M frames for a fair comparison, but neither meaningfully improves after the first 6M, averaging only 5 for PPO and 0.1 for VPN at the end of training (\cref{fig:extended-craft}).

To understand why, consider what is needed to perform well.
If the agent is in a room without any coins in it, the next possible reward may be hundreds of atomic actions away, even with an optimal policy. 
This could correspond to a dozen sequential symbolic steps, or only two sequential ancillary goals. 
As a result, it is a challenging environment for a model-based method like VPN to operate in, even if it could learn a sufficiently accurate model in the first place.
By contrast, SAGE's meta-controller does not need to reason over many steps to perform well. 
We hypothesise that hDQN's comparatively poor performance is due to the fact that constructing essential pieces of equipment such as pickaxes requires precise sequencing of symbolic steps, and a single incorrect action can result in the irreversible waste of the raw components. 

Another advantage of SAGE's planning model is that it enables temporally extended and semantic exploration \cite{nachumWhyDoesHierarchy2019}; it effectively explores in the state space, not the action space.
Consider the difference in the actions chosen by the two methods: for SAGE the meta-controller's actions (ancillary goals) correspond to states of the world it wants to be in: ``I should have a pickaxe."
On the other hand, hDQN is exploring by taking random (albeit high-level) actions: ``face a tree."
This is evidenced by SAGE outperforming hDQN at the start of training, even though both methods are acting randomly with the same low-level controllers.

\section{Experimental reproducibility details}
\label{sec:reproducibility}
All results reported in the paper are generated from 3 random seeds. 
As is standard in the DRL literature, we report the average undiscounted return per episode, which is equal to the number of passengers delivered in the taxi environments and the number of coins collected in CraftWorld.
The shaded area on the graphs represents the bootstrapped 95\% confidence interval.
Experiments were run on Ubuntu 16.04 on a machine with an Intel Core i7 7820X (8 cores), 3 Nvidia RTX 2080Ti GPUs and 64GB of RAM.
With the exception of VPN, for which we used the author's original Tensorflow implementation, we used PyTorch to develop our methods.
Our code is publicly available in the following repository: \url{https://github.com/AndrewPaulChester/sage-code}.

\subsection{PPO}
Our PPO implementation used in both craft and taxi experiments is based on Ilya Kostrikov's \cite{pytorchrl} (\url{https://github.com/ikostrikov/pytorch-a2c-ppo-acktr-gail}).
Unless otherwise stated, we used the suggested hyperparameters for Atari environments as they are the closest to our domains (in terms of observation space, action space and episode lengths).
128-step returns are used, with Generalised Advantage Estimation ($\lambda = 0.95$). 
The value loss coefficient is 0.5, and the PPO clip parameter is set to $0.1$. 
We use 16 actors in all experiments, with 4 PPO epochs and a minibatch size of 4.

\subsection{DQN}
Our DQN implementation used for the meta-controller of the craft experiments is based on rlkit (\url{https://github.com/vitchyr/rlkit}).
With the exceptions of the hyperparameters outlined in \Cref{tab:hyperparameters}, we left the settings at their default values:
\begin{itemize}
    \item replay buffer size: $10^5$
    \item soft target updating each step with $\tau = 10^{-3}$
    \item batch size: 32
    \item The exploration rate ($\epsilon$) starts at 1 and is linearly annealed to 0.1 over the first 200 epochs (600k frames).
    \item Double DQN is used to mitigate overly optimistic Q estimates.
\end{itemize}

\begin{table*}
    \caption{Hyperparameter Details}
    \begin{center}
        \begin{tabular}{c |c |c c c}
        & Taxi & \multicolumn{3}{c}{CraftWorld}\\
        Parameter &PPO and SAGE &  PPO & SAGE low-level & SAGE meta-controller\\\hline
         $\gamma$ (discount rate) & $\mathbf{0.99}$& $\mathbf{0.99}$ & $\mathbf{0.99}$& $\{0.99,\mathbf{0.95},0.9\}$\\ 
         $\alpha$ (learning rate) & $\{2.5e^{-3},1e^{-3},\mathbf{2.5e^{-4}}\}$&  $\{3e^{-3},1e^{-3},\mathbf{3e^{-4}}\}$ & $\{3e^{-3},\mathbf{1e^{-3}},3e^{-4}\}$  &$\{1e^{-3},3e^{-4},1e^{-4},\mathbf{3e^{-5}}\}$\\  
         Adam $\epsilon$ & $\{1e^{-5},\mathbf{1e^{-7}}\}$ &  $\{\mathbf{1e^{-5}},1e^{-7}\}$ & $\{\mathbf{1e^{-5}},1e^{-7}\}$ &  $\{\mathbf{1e^{-5}},1e^{-7}\}$\\  
         PPO entropy & \multirow{2}{*}{$\{0.02,\mathbf{0.006},0.002\}$} & \multirow{2}{*}{$\{\mathbf{0.02},0.006,0.002\}$} &\multirow{2}{*}{$\{\mathbf{0.02},0.006,0.002\}$}& \multirow{2}{*}{N/A}\\  
         coefficient&&&&\\
        \end{tabular}
    \end{center}
    \label{tab:hyperparameters}
    \end{table*}

\subsection{Neural network architectures}
For the PPO, SAGE and hDQN methods across both domains, we have used two different neural network architectures: one convolutional for image input, and one fully connected where images are not used.

The convolutional architecture matches the original DQN implementation \cite{mnihHumanlevelControlDeep2015}, and consists of the following sequential layers:
\begin{itemize}
    \item Conv2D(32 channels, (8, 8) kernel, stride 4)
    \item Conv2D(64 channels, (4, 4) kernel, stride 2)
    \item Conv2D(32 channels, (3, 3) kernel, stride 1)
    \item Dense(512 neurons)
\end{itemize}

The fully connected architecture consists of the following sequential layers:
\begin{itemize}
    \item Dense(256 neurons)
    \item Dense(256 neurons)
\end{itemize}

All hidden layers use ReLu activation, and the final output layer is determined by the action space of the domain and the algorithm used (DQN: $|\mathcal{A}|$ output neurons, PPO: one output neuron for the critic, and one for each binary or continuous variable in the output space.)
The Adam optimiser was used across all algorithms.

\subsection{Experiment hyperparameters and search strategy}
The range of hyperparameters tested for each domain is shown in \Cref{tab:hyperparameters}, with the final values chosen in bold.
We took parameters based on their average score after training for 1M frames for taxi and 6M frames for CraftWorld. Tuning was not performed independently for each size of taxi environment - parameters were set on the small environment (as that had the largest signal for all methods) and then applied across the other sizes.
Parameters for the craft low-level controller and meta-controller were set independently as they use different algorithms (PPO and DQN respectively).
As it is an ablation, hDQN used the same parameters as SAGE.

\subsection{VPN}
For our VPN benchmarks we used the authors' original code (\url{https://github.com/junhyukoh/value-prediction-network}). 
As our environments are largely comparable to standard Atari domains, we used their settings from the paper for the Atari experiments.
The only hyperparameters changed were the prediction step, which we increased to 10 (from 5 in the original paper) as our environments had sparser rewards and therefore required a longer lookahead.

\subsection{SDRL}
The SDRL benchmarks was recoded from scratch according to the details in the original paper and can be found in the code appendix. 
As our environments are almost identical to their taxi environment, we kept the hyperparameters from the original PEORL paper \cite{yangPEORLIntegratingSymbolic2018}, with $\beta=0.5$ and $\alpha$ linearly annealed from 1 to 0.01 over the first 200 episodes of training.
Since our taxi environment did not require low-level control, there was no neural network involved, and the hyperparameters for subtask completion are not used.

\section{Conclusion}
In this paper we have introduced SAGE, a hierarchal agent combining deep reinforcement learning and symbolic planning approaches. 
This neuro-symbolic architecture allows us to exploit easily defined explicit knowledge about the environment while maintaining the flexibility of learning unknown dynamics. 
SAGE outperforms both planning and learning based approaches on complex problems, particularly as the environments are scaled up. 
To the best of our knowledge, this is the first work in which an RL agent sets symbolic goals for a planner independently, rather than simply selecting between a handful of predetermined templates. 

A limitation of our method is that all actions must go through the planner, which requires the provided model to address all aspects of the domain. 
In Real Time Strategy games like StarCraft II, building and unit production is nicely captured by a myopic and abstract model, but combat is not. 
An interesting line of future work would be to allow the meta-controller to selectively bypass the planner, directly setting ancillary goals for the low-level controller when appropriate.

Another limitation is that the agent is committed to completing the full plan at the outset; there is no opportunity for the meta-controller to `change its mind' unless a step times out. 
This could be addressed by replanning at fixed intervals or using a more sophisticated approach that detects the presence of new or unexpected information and shifts control back to the meta-controller \cite{ahaGoalReasoningFoundations2018}.

In the future, we would also like to extend this work with tighter feedback between the planner and the meta-controller, and try to learn updates to the provided PDDL model \cite{ngIncrementalLearningPlanning2019}.

\ifCLASSOPTIONcaptionsoff
  \newpage
\fi



\bibliographystyle{IEEEtran}
\bibliography{symbolic-goal-generation}
%



%

\begin{IEEEbiographynophoto}{Andrew Chester}
Biography text to be added upon acceptance.
\end{IEEEbiographynophoto}

\begin{IEEEbiographynophoto}{Michael Dann}
    Biography text to be added upon acceptance.
\end{IEEEbiographynophoto}


\begin{IEEEbiographynophoto}{Fabio Zambetta}
    Biography text to be added upon acceptance.
\end{IEEEbiographynophoto}

\begin{IEEEbiographynophoto}{John Thangarajah}
    Biography text to be added upon acceptance.
\end{IEEEbiographynophoto}



\end{document}